# Beyond Scalar Objectives: Expert-Feedback-Driven Autonomous Experimentation for Scientific Discovery at the Nanoscale

Ralph Bulanadi[1], Jefferey Baxter[1], Arpan Biswas[2], Hiroshi Funakubo[3], Dennis Meier[4,5,6], Jan Schultheiß[4], Rama Vasudevan[1], Yongtao Liu[1*]

[1]Center for Nanophase Materials Sciences, Oak Ridge National Laboratory, Oak Ridge, TN 37831, USA

[2]University of Tennessee-Oak Ridge Innovation Institute, University of Tennessee, Knoxville, TN 37996, USA

[3]Department of Material Science and Engineering, School of Materials and Chemical Technology, Institute of Science Tokyo, Yokohama, 226-8502, Japan

[4]Department of Materials Science and Engineering, Norwegian University of Science and Technology (NTNU), Trondheim, Norway.

[5]Faculty of Physics and Center for Nanointegration Duisburg-Essen (CENIDE), University of Duisburg-Essen, Duisburg, Germany.

[6]Research Center Future Energy Materials and Systems, Research Alliance Ruhr, 44780 Bochum, Germany

*liuy3@ornl.gov

**Abstract**

Self-driving laboratories or autonomous experimentation are emerging as transformative platforms for accelerating scientific discovery. Bayesian optimization (BO) is among the most widely used machine learning frameworks in self-driving laboratories, but these BO based frameworks rely on predefined scalar descriptors to guide experimentation. This limits their ability to handle complex phenomena that are difficult to formalize as scalar descriptors. In many situations, the determination of an appropriate scalar descriptor can be challenging, and may fail to capture subtle yet scientifically important phenomena apparent to experts with interdisciplinary insight. This problem is particularly pronounced in nanoscale imaging experiments, where strong local heterogeneity, multidimensional observations, and emergent physical behaviors often cannot be effectively represented using scalar metrics. To overcome this limitation, here we develop deep-kernel pairwise learning (DKPL), an approach for autonomous microscopy experiments which incorporates human expertise and interdisciplinary scientific knowledge into an active learning loop. Instead of relying on explicit scalar objectives, DKPL enables experts to directly evaluate which experimental output is more promising using interdisciplinary knowledge. DKPL then learns a latent utility function from these expert judgements to guide subsequent autonomous microscopy experiments. We demonstrate DKPL's performance in learning physically meaningful nanoscale structures while effectively prioritizing high-information measurement regions using an experimental model dataset with known ground truth. We further apply DKPL to analyze the character of ferroelectric domain walls, where we find DKPL capable of distinguishing between high and low characteristic domain-wall angles in bismuth ferrite, and able to discover both head-to-head and tail-to-tail domain-wall character in erbium manganite. This development establishes an approach to integrate expert knowledge into autonomous microscopy experiments and demonstrates a pathway toward expert-guided self-driving laboratories capable of addressing scientific problems beyond the limits of scalar-metrics-driven learning.

**Keywords:** Self-Driving Laboratory; Automated and Autonomous Experiments; Microscopy; Human-AI Collaboration; 3D Polarization Mapping; Ferroelectrics; AI/ML

## Introduction

Autonomous experimentation (AE) or self-driving laboratories (SDL) represent a transformative paradigm in materials science research[1-5] which integrate robotic systems with artificial intelligence (AI) and machine learning (ML) to conduct experiments autonomously. For materials scientists, AE can explore vast compositional and processing parameter spaces efficiently, running hundreds or thousands of experiments in a short time frame that would not be possible through traditional manual approaches. AE has demonstrated remarkable applications in optimizing batteries[6], catalysts[7], organic molecules[8], functional ceramics[9], 2D materials[10,11], and thin films for ferroelectrics and photovoltaics[12-14]. These platforms typically employ active learning strategies that intelligently select subsequent experiment parameters based on accumulated data, dramatically accelerating optimal materials design.

Active learning workflows that are commonly used in AE, such as Bayesian optimization, rely on numerical values to guide the experimental process. These numerical values are often not directly measurable, but instead are derived from raw experimental data, such as imaging or spectroscopy which are reduced to a scalar descriptor or performance metric (e.g., peak intensity, band gap, or grain size) for the optimization algorithm. These quantitative, scalar metrics act as the feedback signal, closing the loop between experimental execution, data analysis, and experiment design.

However, there are inherent limitations when converting raw experimental results (e.g., spectra, images, or higher dimensional hyperspectral data) into scalar metrics. This conversion typically requires the data analysis or scalar-extraction strategy to be predefined, based on prior knowledge or expectations about the phenomena of interest. In practice, however, it is fundamentally impossible to anticipate all relevant behaviors that may emerge in novel experiments, particularly when AEs conduct hundreds or thousands of experiments across previously unexplored compositions or parameter spaces. As a result, predefined scalar metrics may likely introduce information loss, false optima, or misleading outliers into the active learning process.[15-17] For instance, unexpectedly noisy raw data can lead a peak-finding algorithm (such as a scalar-extraction strategy) to produce meaningless scalar values, yet these erroneous values would still be treated as feedback equally as valid as any meaningful values by the autonomous system for subsequent experiment selection.

In addition, many important physical phenomena cannot be adequately described as numerical values. New or complex phenomena may lack well-defined metrics, particularly in early-stage exploratory research where prior knowledge may be flawed or incomplete. These situations provide ample potential for on-the-fly human feedback [18], where the overall promise of a novel material or phenomenon may be immediately evident to a domain expert from raw data, yet resist a simple scalar metric. Similarly, relevant properties frequently involve subjective judgments or multidimensional trade-offs that cannot be captured by a single number without losing critical information. This fundamental limitation creates a critical gap in AE, where promising research directions that cannot be represented by numerical metrics resist exploration by active learning.

By integrating domain expertise, including interdisciplinary knowledge, pattern recognition, and an assessment of qualitative differences into AE, decisions regarding complicated phenomena could be made such that discovery is not limited to what is already quantifiable.

We address this requirement in this work by developing a preference-driven active learning framework based on a pairwise Gaussian process (GP)[19] that learns from expert comparisons rather than quantitative metrics. Preference-learning methods, including pairwise comparisons [20-22], allow a system to infer priorities from a domain expert's choices between experimental outcomes, rather than relying on numerical values as feedback. This allows AE to be driven by the nuanced judgment of domain experts. To account for high-dimensional input, we also add deep kernel learning, a neural network that first extracts low-dimensional features from high-dimensional inputs, transforming complex data, such as image patches, into a low-dimensional representation. A pairwise GP then operates on this learned feature space to model the latent utility function. This approach enables the system to handle complex relationships while maintaining principled uncertainty estimates essential for active learning. Furthermore, we integrate support for indifference comparisons, where two options are considered equally good, together with confidence weighting for uncertainty in expert judgments. These features make the framework particularly suitable for active learning scenarios where collecting a reliable absolute score is difficult or unreliable. This framework is first tested on well-understood systems with a known ground truth to provide a solid benchmark for evaluating its ability to reconstruct physically relevant domain structures from human feedback. We show that this approach efficiently identifies high-information measurement regions without predefined scalar metrics. We then apply it to autonomous investigation of complex domain-wall structures in ferroelectric thin films, where multidimensional polarization behaviors cannot be adequately represented by scalar descriptors. The framework enables autonomous discovery and prioritization of scientifically relevant polarization configurations guided by real-time human-feedback.

**Deep Kernel Pairwise Learning**

Typically, traditional active learning approach in AE or SDL rely on scalar-valued objective functions, where experimental outcomes (e.g., 1D spectra, 2D images, etc.) are converted to numerical scores that serve as inputs for surrogate models. While this can be straightforward, this paradigm requires the prior definition of metrics or scalar functions that map complex high-dimensional experimental observations to scalar values. However, in many scenarios, such metrics are difficult to define. In addition, in many scientific settings, decision-making is inherently multi-objective. Experts must simultaneously weigh competing or complementary factors when evaluating experimental outcomes, such as performance, stability, and cost. Although these considerations implicitly define an underlying utility function, explicit formalization of this function, with precise scoring and weighting for each objective, is often impractical. In contrast, humans are generally much more reliable when making relative judgments, e.g., given two outcomes, human can readily indicate a preference by leveraging interdisciplinary knowledge, even when the decision reflects multiple interacting criteria.

As such, we consider preference-driven learning based on Pairwise GP as an alternative formulation to handle complicated experimental outcomes or intricated relationships while avoiding explicit scalar quantification. Instead of assigning absolute numerical scores, Pairwise GP learns from pairwise comparisons in which an expert indicates which of two experimental outcomes is more promising (e.g., "A is better than B"). From these comparisons, the model infers a latent utility function that implicitly captures expert judgment. This approach allows optimization toward desirable outcomes without requiring predefined metrics, making it particularly suitable for complex phenomena where expert assessment integrates multiple qualitative and quantitative factors, or where raw data contains rich structure that would be lost through scalar reduction.

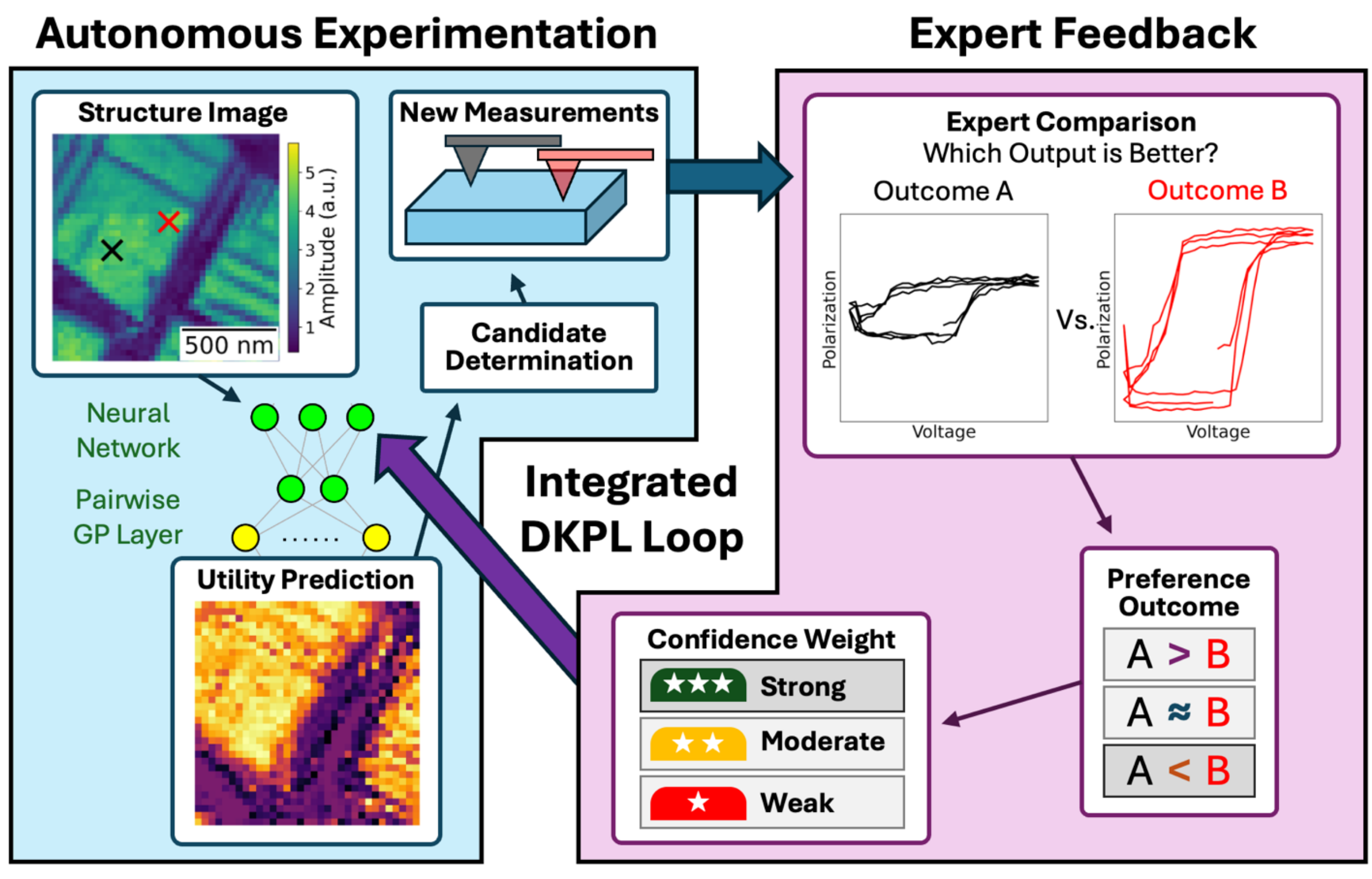

**Figure 1.** The DKPL process, showing the integration of expert feedback into autonomous experimentation.

To integrate preference learning into microscopy experiments, where 2D images are used as inputs, we incorporate a pairwise GP model with a deep-kernel learning framework: a process we name deep kernel pairwise learning (DKPL). In DKPL, as shown in Fig. 1, a neural network serves as a learnable feature extractor, mapping raw high-dimensional data (e.g., microscopy images), into a lower-dimensional latent representation. The pairwise GP operates on the latent representation to map the relationship between high-dimensional data (e.g., image patches) and expert's preference on the related experimental observation (e.g., spectroscopy measurement). The GP layer further provides principled uncertainty quantification, which is essential for active learning strategies that prioritize informative comparisons.

The complete model is trained via joint gradient-based optimization of the neural network parameters and the pairwise GP hyperparameters by maximizing the marginal likelihood. In detail, after random sampling, two new points are selected through an upper confidence bound (UCB) strategy (by default, $\beta = 5$). These two points are compared with one another, and each of these two points are also compared to the current highest-utility point. These three comparisons are then used to predict a new utility map over 1000 epochs, and the process repeats.

To improve performance in experimental settings, we extend the pairwise comparison to accommodate more nuanced human feedback. First, we support equal-preference judgments: rather than force a strict ordering between outcomes, we introduce a tolerance range in which two outcomes are treated as equally preferred if their latent utilities differ by less than a predefined threshold. This mechanism reduces forced-choice bias and more accurately reflects expert judgment, especially when many candidates appear similarly promising. Second, we incorporate confidence-weighted comparisons to account for uncertainty in human decisions. Experts may occasionally be unable to make confident judgments due to noisy observations or subtle differences between outcomes. Each comparison can therefore be assigned a confidence (weak, moderate, or strong, which map to a numerical training weight) reflecting the reliability of expert judgment. During training, high-confidence comparisons exert greater influence on parameter updates, while low-confidence judgments contribute proportionally less. This weighting scheme allows the model to learn from uncertain feedback without allowing it to dominate training and enables the system to capture patterns in user reliability over time. Together, these components form DKPL enabling preference-driven active learning capable of operating directly on rich, high-dimensional experimental data with expert judgment.

**DKPL in Band Excitation Piezoresponse Spectroscopy**

We implemented DKPL on pre-acquired band excitation piezoresponse spectroscopy (BEPS) data of a lead titanate ($PbTiO_3$) thin film. Band excitation piezoresponse image and spectroscopy measurements each provide detailed characterization of ferroelectric materials[23,24]. Images of piezoresponse amplitude maps (labelled “Structure Image” in Fig. 1) and phase-offset maps reveal distinct domain structures: *a*-domains with in-plane polarization and *c*-domains with out-of-plane polarization. Spectroscopy measurements, on the other hand, can probe local switching behavior and polarization dynamics by measuring piezoresponse as a function of applied DC voltage, generating hysteresis loops that capture critical ferroelectric characteristics, such as coercive voltage, nucleation behavior, and saturation polarization [25]. Understanding the relationship between these underlying phenomena (that is, how local domain structures from images influence spectroscopy data) is fundamental to establishing structure–property relationships in ferroelectric materials. Such relationships reveal, for example, how domains, structural defects, and other nanoscale features affect polarization switching mechanisms [26,27].

Our prior work introduced an active learning framework based on deep kernel learning (DKL) that automated the exploration of structure–property relationships[28]. This approach used scalar descriptors extracted from spectroscopic data, such as hysteresis loops[28,29], current–voltage curves[30], nonlinearity spectra[31], or novelty of spectra[32], to guide spectroscopy measurement selection. However, this scalar-based DKL approach had inherent limitations. In this case, the scalar-based approach condenses complex spectroscopy data into single numerical descriptors and target parameters, rather than establishing explicit connections between spatial image features and their holistic spectroscopic signatures. DKPL addresses these shortcomings directly, indicating promising experimental results to guide subsequent measurements. This enables autonomous exploration of complex phenomena that defy reduction to scalar metrics, effectively combining the pattern recognition capabilities of human experts with the efficiency of machine-driven experimentation.

To verify the validity of DKPL to translate expert selection into spatial predictions, compared the utility maps based on calculated hysteresis-loop-area size against the hysteresis-loop-area ground truth (Fig. 2a). The utility here represents learned relative preferences based on judgement of the hysteresis loop area, rather than an absolute scalar metric. We consider loop area here as a proxy for understanding, with larger loop areas often indicating more meaningful characteristics. Furthermore, the loop area map reveals domain structures, with high-value regions (lighter regions) corresponding to out-of-plane *c*-domains, delineated by darker stripe features corresponding to in-plane *a*-domains with characteristically smaller loop opening.

It is important to clarify that the utility values predicted by the model do not directly correspond to loop area values. If point A has utility 0.5 and point B has utility 0.1, this only indicates that A exhibits larger loop opening than B, but not any other numerical relationship between the two points.

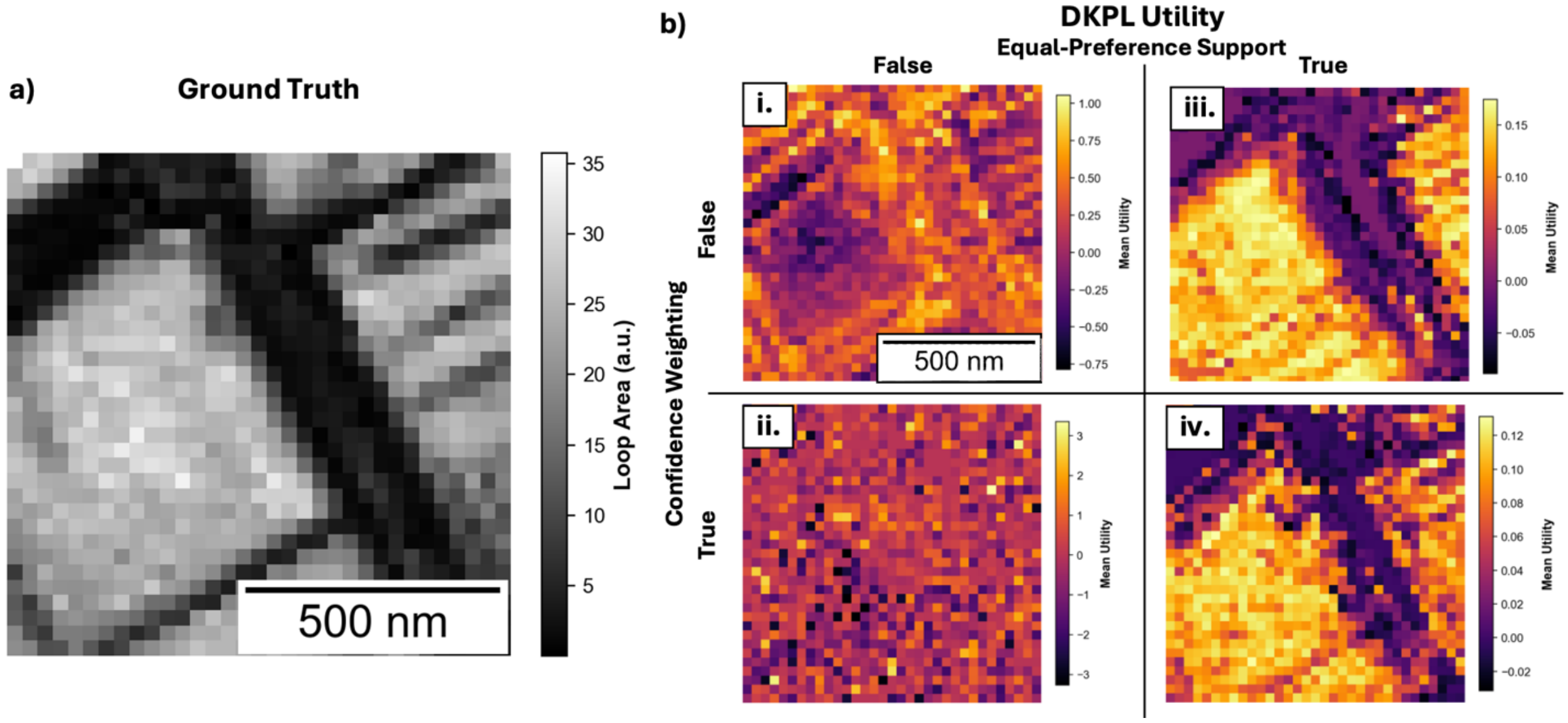

**Figure 2.** DKPL predictions in BEPS experiments. a) The loop-area "ground truth" calculated from spectroscopic data. b) DKPL utility predictions for i. reference DKPL, with neither confidence weighting nor equal-preference support; ii. confidence weighting but not equal-preference support; iii. equal-preference support but not confidence weighting; and iv. both equal-preference support and confidence weighting.

The initial DKPL model, which does not implement confidence weighting or equal-preference support, generates noisy predictions devoid of recognizable domain features (Fig. 2b.i.). Incorporating the confidence weight alone does not immediately lead to notable improvement in predicting domain structures (Fig. 2b.ii.). However, equal-preference support alone leads to notable improvement in predicting spatial domain structures (Fig. 2b.iii.). Incorporating both confidence weighting and equal-tie support also captures the domain structure, leading to predictions with sharper domain contrast and superior differentiation between high and low utility regions (Fig. 2b.iv.). This ability to reconstruct complex domain structures from sparse pairwise comparisons demonstrates that the DKPL model indeed learns physical correlations between nanoscale structures based on expert selection. For autonomous exploration, this enables DKPL to not only identify where interesting spectroscopic features occur, but also identify domain structures in which such phenomena are likely to manifest.

We further explored how our DKPL implementation affects sampling priorities, which in turn determines experimental efficiency and scientific yield. Figure 3 reveals how each implementation method reshapes exploration behavior through spatial sampling maps and corresponding loop area distributions. With neither confidence weighting nor equal-preference support, the algorithm concentrates measurements toward in-plane *a*-domains, where BEPS hysteresis measurements yield poor signal-to-noise ratios (Fig. 3a.i.). The loop area histogram confirms this problematic bias with a low-value peak (normalized to 0.0–0.1) (Fig. 3a.ii.). This represents a critical failure mode when treating all pairwise inputs as equally important (particularly treating comparisons between two informative open loops as equally important as comparisons between two uninformative closed loops) causes wasteful sampling in uninformative regions. The addition of confidence weighting produces scattered and more uniform sampling throughout the sample (Fig. 3b). This, combined with the generally uniform utility predictions shown in Fig. 2b.ii. further suggests that confidence-weighting alone does allow for meaningful utility calculations. Equal-preference support, however, leads to sampling focused on two *c*-domain regions (Fig. 3c.i.), with loop area distribution with a peak at high (0.6–0.8) values (Fig. 3c.ii.). The loop area distribution shifts away from extreme low values, indicating the model has learned to deprioritize unreliable spectroscopic responses and, as such, equal-preference support helps prioritize high-utility regions. Combining confidence weighting with equal-preference support demonstrates even higher preference sampling in regions corresponding to *c*-domains (Fig. 3d) compared to equal-preference support alone. This focused pattern indicates the model has successfully learned both spatial locations of interesting phenomena and informative spectroscopic signatures.

.

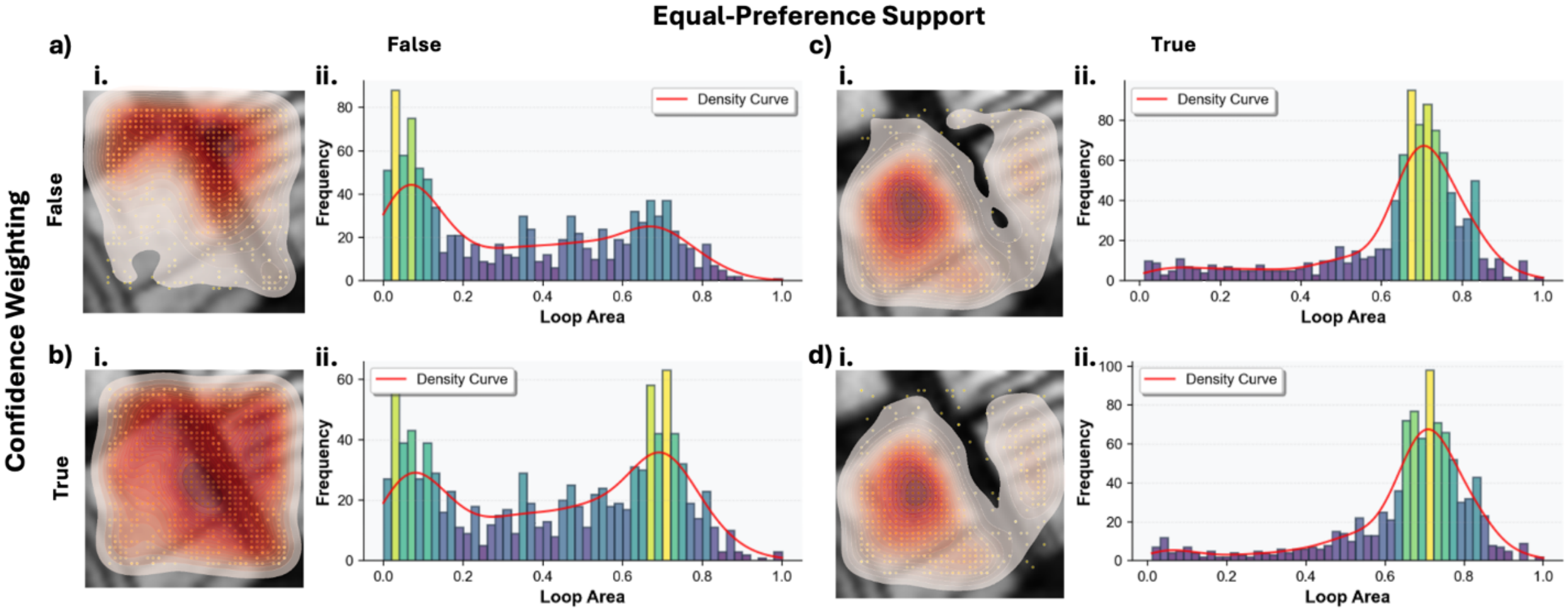

**Figure 3.** DKPL sampling in BEPS experiments. a) Reference DKPL, with neither confidence weighting nor equal-preference support; b). confidence weighting but not equal-preference support; c) equal-preference support but not confidence weighting; and d). both equal-preference support and confidence weighting. For each, i. shows a distribution of sampling points, and ii. Shows a histogram map of measured loop areas.

Altogether, equal-preference support appears to be a crucial element for pairwise comparisons, enabling both more realistic reconstructions of the underlying dataset, and improved prioritization of high-information regions. The addition of confidence weighting alone, which appears to penalize over-sampling in low signal-to-noise regions. Without equal-preference support, this deprioritization leads confidence weighting alone to fail at a meaningful reconstruction, but the combination of confidence weighting with equal-preference support yields subtle improvements in both sampling and reconstruction than equal-preference support alone. As such, the combination of both confidence weighting with equal-preference support promotes meaningful exploration by identifying regions where predictions show competing alternatives, guiding strategic exploration toward spatially coherent features of scientific interest and physically meaningful structures.

**DKPL to determine the characteristic orientation angles of domain walls**

We further leverage DKPL to analyze three-dimensional polarization vectors as measured by Auto-3DPFM[33]. In this method, the three-dimensional polarization vector of a sample is reconstructed from multiple piezoresponse force microscopy (PFM) images, which provides a map of a sample's three-dimensional polarization state, but the resulting analysis can be unwieldy to examine analytically. These problems are emphasized in studies of ferroelectric domain walls, which often have fascinating emergent physics determined by the characteristic angle between the polarization vectors of neighboring domains[34-37]—this characteristic angle is defined by spatial (two-dimensional) relationships of local three-dimensional phasors. While methods have been developed to numerically calculate the characteristic angle of a domain wall [33], this is computationally expensive and must be recalculated for every pixel of every image; meanwhile, human experts can develop an understanding of these high-dimensional relationships.

An example of such an Auto-3DPFM result on a rhombohedral, polycrystalline lead zirconate titanate (PZT) sample is shown in Fig. 4a. Due to electrostatic and mechanical boundary conditions, domain walls in this sample have a characteristic angle of 71°, 109°, or 180°. In Fig. 4.a.ii, the local polarization vectors change by a lower angle (~71°), while in Fig. 4.a.iii, the local polarization vectors change dramatically (~180°), which can immediately be perceived as an interface between two regions of similar polarization directions. This presents a key use case for DKPL. The model, given sufficient input by a human expert, could be driven to differentiate domain walls by their characteristic angle.

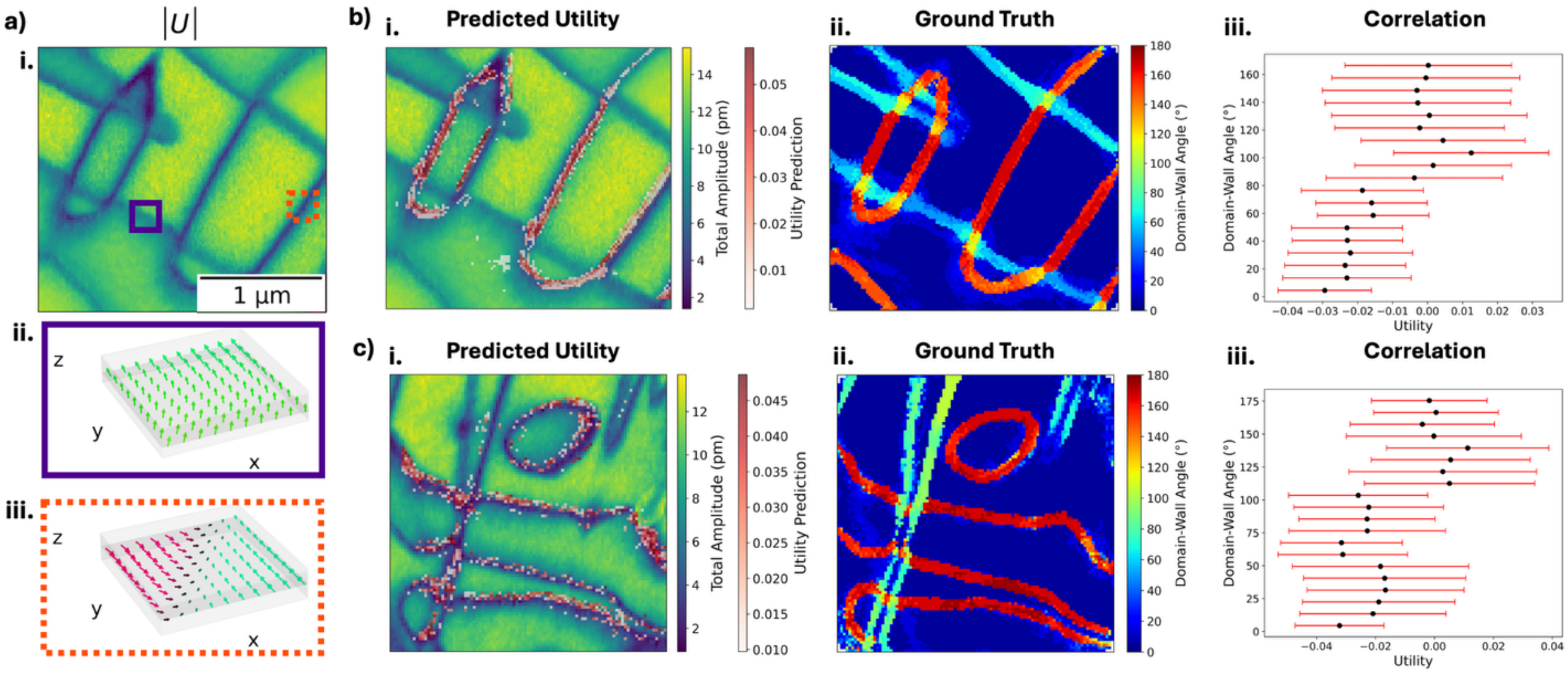

**Figure 4.** DKPL sampling in Auto-3DPFM of PZT to examine characteristic domain-wall angle. a) i. Map of amplitude. The solid purple square shows (ii.) a domain wall without 180° character; the dotted orange square shows (iii.) a domain wall with 180° character. b) The results of DKPL, showing i. the highest-predicted-utility points overlayed on the total amplitude; ii. the "ground truth" domain-wall character established through the most-common vector-angle difference method, and iii. The correlation between predicted utility and the measured domain wall angle. For c), the model trained in b) was directly used on a distinct region of the sample, once more showing i. predicted utility, ii. ground truth, and iii. correlations.

In this investigation, the DKPL was input with human expertise to identify 180° domain walls in a sample of rhombohedral polycrystalline PZT. Additional pairs were selected 20 times, with comparisons made between points of maximum UCB ($\beta$ = 10000 for steps 1 through 10; and $\beta$ = 2 for steps 11 through 20) and maximum uncertainty. The regions with the highest utility are plotted in Fig. 4b.i., overlaid onto a map of the total amplitude. This utility can be compared to the "ground truth" domain-wall angle determined by the most-common angle difference method (Fig. 4b.ii.); these high-utility regions can indeed be seen to correlate with 180° domain walls. These can be verified in Fig. 4.b.iii., which shows the distribution of utilities against the domain wall's true characteristic angle; a dramatic increase in utility can be seen for domain walls of angle above 90°.

It is also notable that the "domain-wall crossings" between the 71° and 180° domain walls are given an intermediate characteristic angle in the "ground-truth" mode calculation which is built using scalar metrics. This method, as outlined in Ref. [33], calculates the vector-angle differences between pairs of points positioned symmetrically opposite an arbitrary point, and assigns the characteristic angle of that arbitrary point to be the most-common vector-angle difference from all such pairs within a radius of that point. While this method works particularly well across most domains or domain walls, this system fails at the domain-wall crossings, where multiple characteristic angles may be equally valid, while the most-common vector-angle difference must be only one particular angle, or be an intermediate angle. In contrast, when DKPL is trained by human expertise to find 180° domain walls, it correctly assigns a high utility value to crossings that contain a 180° domain wall. The DKPL model, in directly integrating human judgement, may thus represent a *more* truthful and meaningful measurement of the domain wall character than the mode calculation.

To verify that the model was not simply overtrained on a particular region, this trained model was used on another region of the same sample, with the utility overlay, "ground truth", and utility-angle correlation shown in Fig. 4c. Once more, the utility appears to overlay specifically with the 180° domain walls, suggesting the strong capability of this model to determine 180° domain walls.

**DKPL to identify the polar character of domain walls**

While this model was successful in differentiating 180° domain walls from non-180° domain walls, this distinction is primarily categorical; a region either corresponds to a 180° domain wall, or it does not. To example the capability for DKPL to study *continuous* phenomena, we instead move to studies of polycrystalline erbium manganite ($ErMnO_3$)[38]. These samples present curved domain walls, where the charge state of head-to-head or tail-to-tail walls continuously varies as the angle of the domain wall with respect to the polar axis changes[39,40]. The corresponding bound charges at head-to-head or tail-to-tail walls have been shown to determine the local conduction in $ErMnO_3$, due to the buildup of mobile charge carriers. Once more, this is an extensive, high-dimensional optimization problem, but one with a simple question for human-expert pairwise comparison: which domain wall has more "head-to-head" and which has more "tail-to-tail" character?

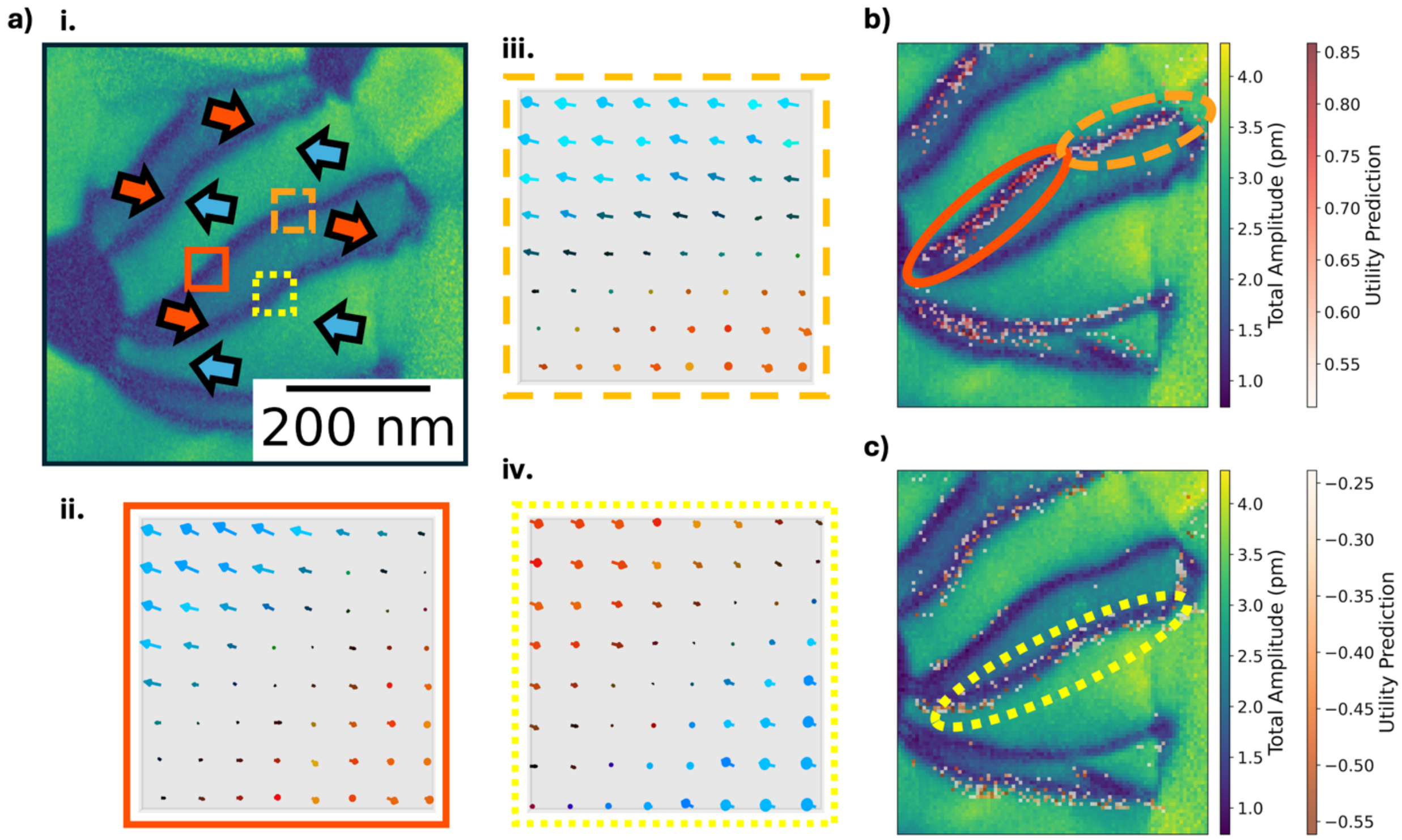

**Figure 5.** DKPL sampling in Auto-3DPFM of a grain in an $ErMnO_3$ polycrystal to examine the tail-to-tail character of domain walls. a) i. Map of amplitude with ferroelectric domains marked. The solid red square shows (ii.) strong tail-to-tail character; the dashed orange square shows (iii.) weak tail-to-tail character; and the dotted yellow square shows (iv.) head-to-head character. b) An overlay showing highest-predicted-utility points highlights tail-to-tail character, with regions of stronger tail-to-tail character generally showing higher utility (solid red ellipse vs. dashed orange ellipse). c) An overlay showing lowest-predicted-utility points highlights head-to-head character (dotted yellow ellipse).

Auto-3DPFM data of a region of $ErMnO_3$ is shown in Fig. 5a; here, all domain walls are 180° walls. Some regions maintain a tail-to-tail character (ii, iii), while others maintain a head-to-head character (iv). Of those that are in tail-to-tail configuration, due to the orientation of the wall itself, regions can be "more" or "less" tail-to-tail than others. Due to possessing such a continuous character, more strict comparisons can be made: For the model generated for PZT, 56% of comparisons were equal ties, while for the model generated $ErMnO_3$, 18% of comparisons were equal ties. This results in much higher values for model utility and improved differentiation between regions.

An overlay of regions with the highest utility, when trained by a human expert to prioritize regions of maximal tail-to-tail character, are presented in Fig. 5b. Here, only the tail-to-tail domain walls are highlighted. In addition to this, the regions with the highest utility (marked in red) tend to segregate along the more vertical domain walls, showing the capability of this model to differentiate between continuous phenomena.

An emergent property of this training method, while it is not the main goal of this method, is that the utility function also identifies regions with the *least* head-to-head character (and thus, the most tail-to-tail character). The same model, with the regions of lowest utility highlighted, are instead shown in Fig. 5c. Here, the opposite domain walls are highlighted. Unlike the cases for tail-to-tail domain walls, there does not appear to be differentiation between how much of a head-to-head character a domain wall is.

**Conclusions**

In summary, we developed DKPL, an expert-feedback-driven framework that enables the integration of analytically complex yet heuristic decision making into autonomous experiments. DKPL combines pairwise preference learning, confidence-weighted expert feedback, and support for indifference comparisons, to learn a latent utility function that guides autonomous exploration without requiring predefined scalar objectives. Unlike commonly used BO workflows that depend on explicit numerical metrics as feedback, DKPL incorporates implicit domain knowledge and interdisciplinary scientific expertise directly into the experimental decision-making process. Through autonomous scanning probe microscopy studies, we demonstrate that DKPL can identify scientifically relevant phenomena, prioritize high-information measurements, and effectively guide discovery of multidimensional physical behavior. More broadly, DKPL establishes a strategy for expert-guided self-driving laboratories by leveraging real-time human insight in situations where traditional scalar metrics driven decision making is insufficient. This capability is particularly important for scientific problems where meaningful physical phenomena cannot be adequately represented using scalar metrics. We believe this approach will provide a foundation for future autonomous experimentation platforms capable of addressing increasingly complex and poorly formalized discovery problems across materials science and beyond.

**Author Contributions:**

Y.L. conceived the idea. R.B. and Y.L. conducted the investigation. A.B. conducted preliminary investigation with the pairwise GP model. H.F. provided the lead titanate sample. D.M. and J.S. provided the polycrystalline erbium manganite sample; J.B. performed additional post-processing. R.B. and Y.L. wrote the manuscript. All authors edited the manuscript.

**Acknowledgements:**

The authors thank Dr. Kyle Kelley for fruitful discussions regarding sample preparation. This research and workflow development was sponsored by the INTERSECT Initiative as part of the Laboratory Directed Research and Development Program of Oak Ridge National Laboratory, managed by UT-Battelle, LLC for the US Department of Energy under contract DE-AC05-00OR22725. Piezoresponse force microscopy was performed at and supported by the Center for Nanophase Materials Sciences (CNMS), which is a US Department of Energy, Office of Science User Facility at Oak Ridge National Laboratory. A.B. acknowledges the use of facilities and instrumentation at the UT Knoxville Institute for Advanced Materials and Manufacturing (IAMM) and the Shull Wollan Center (SWC) supported in part by the National Science Foundation Materials Research Science and Engineering Center program through the UT Knoxville Center for Advanced Materials and Manufacturing (DMR-2309083). H.F. was supported by MEXT Initiative to Establish Next-generation Novel Integrated Circuits Centers (X-NICS) (JPJ011438), and the Japan Science and Technology Agency (JST) as part of Adopting Sustainable Partnerships for Innovative Research Ecosystem (ASPIRE), Grant Number JPMJAP2312. D.M. and J.S. receive funding from the European Research Council (ERC) under the European Union's Horizon 2020 research and innovation program (Grant Agreement No. 863691).

**Conflicts of Interest:**

The authors declare no conflict of interest.

**Data and Code Availability**

The data and code presented in this work is available on GitHub (https://github.com/rbulanadi/DeepKernelPairwiseLearning) [41].